\pdfoutput=1

\documentclass[11pt]{article}

\usepackage[final]{acl}

\usepackage{times}
\usepackage{latexsym}

\usepackage[T1]{fontenc}

\usepackage[utf8]{inputenc}

\usepackage{microtype}

\usepackage{inconsolata}

\usepackage{enumitem}
\setitemize{noitemsep, topsep=0pt, parsep=0pt, partopsep=0pt}

\usepackage{dsfont}
\usepackage{amsmath, amssymb}
\usepackage{graphicx}

\usepackage{booktabs}

\newcommand{\tabincell}[2]{\begin{tabular}{@{}#1@{}}#2\end{tabular}}
\usepackage{multirow}

\usepackage{arydshln}
\usepackage{fancyvrb}
\usepackage{fvextra}

\title{Concise and Precise Context Compression for Tool-Using Language Models}

\author{
	Yang Xu$^{1}$\thanks{work done during internship at Huawei},
	Yunlong Feng$^{1}$,
	Honglin Mu$^{1}$,
	Yutai Hou$^{2}$,
	Yitong Li$^{2}$,
	Xinghao Wang$^{2}$, \\
	{\bf Wanjun Zhong$^{2}$,
	Zhongyang Li$^{2}$,
	Dandan Tu$^{2}$,
	Qingfu Zhu$^{1}$\thanks{corresponding author},
	Min Zhang$^{1}$,
	Wanxiang Che$^{1}$} \\
	$^{1}$Harbin Institute of Technology \\
	$^{2}$Huawei Technologies Co., Ltd \\
	\texttt{\{yxu,qfzhu\}@ir.hit.edu.cn}
}

\begin{document}
\maketitle
\begin{abstract}

Through reading the documentation in the context, tool-using language models can dynamically extend their capability using external tools.
The cost is that we have to input lengthy documentation every time the model needs to use the tool, occupying the input window as well as slowing down the decoding process.
Given the progress in general-purpose compression, soft context compression is a suitable approach to alleviate the problem.
However, when compressing tool documentation, existing methods suffer from the weaknesses of key information loss (specifically, tool/parameter name errors) and difficulty in adjusting the length of compressed sequences based on documentation lengths.
To address these problems, we propose two strategies for compressing tool documentation into concise and precise summary sequences for tool-using language models.
1) Selective compression strategy mitigates key information loss by deliberately retaining key information as raw text tokens.
2) Block compression strategy involves dividing tool documentation into short chunks and then employing a fixed-length compression model to achieve variable-length compression. This strategy facilitates the flexible adjustment of the compression ratio.
Results on API-Bank and APIBench show that our approach reaches a performance comparable to the upper-bound baseline under up to 16x compression ratio.

\end{abstract}

\section{Introduction}

The advent of tool-using language models represents a significant extension of the capability of language models, including but not limited to interacting with the Internet through the web browser, retrieving knowledge from an extended database, or even driving other models or devices~\citep{nakano2021webgpt, OpenAI_2023, patil2023gorilla, cheng2023binding}.
As a common practice, a tool-using language model can be modeled as a function calling model built upon the natural language interface~\citep{patil2023gorilla, li2023api, qin2023toolllm, tang2023toolalpaca}.
Specifically, the user inputs the tool documentation along with the instruction, then the model generates a structured function call defined by the documentation.

\begin{figure}[t!]
	\centering
	\includegraphics{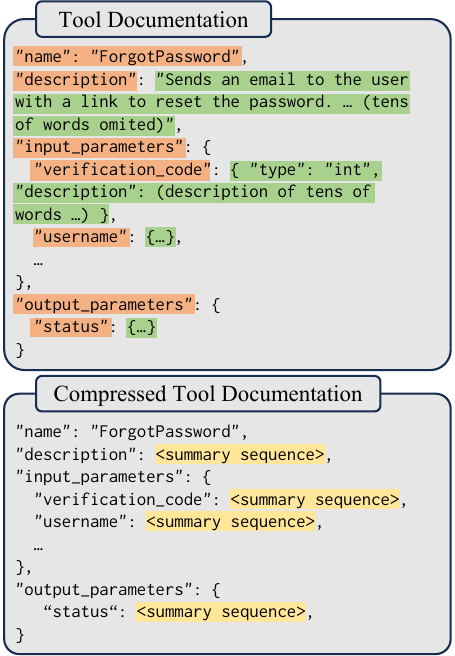}
	\caption{
		An example of tool documentation compression.
		Compared with the key information (red), the other content (green) is more verbose and suitable for compression into summary sequences (yellow).
	}
	\label{fig:task}
\end{figure}

Nevertheless, the tool documentation can be lengthy, occupying a fixed-size input window as long as we use the tool.
To fully unleash the capability of tool-using language models, the tool documentation has to include a detailed description of the functionality and input/output format, which can cost hundreds of tokens per tool.
Moreover, in the cases of multiple tools, like the combination of tools or online top-k retrieval of suitable tools, documentation can easily accumulate to over a thousand tokens.

Fortunately, as shown in Figure~\ref{fig:task}, tool documentation shows the potential to be compressed given its nature.
First, the model only needs to understand the tool's functionality no matter how it is described, which means that we can use a summarized version for most parts of the lengthy documentation.
In contrast, the parts containing key information, such as names and formats marked red in Figure~\ref{fig:task}, must be kept as is, or the generated function call can easily fail. In this work, we regard the key information as names of tools and parameters in the tool documentation.
Second, the documentation is fixed once the tool is deployed, enabling a one-time compression as pre-processing.

In this work, we propose to compress the tool documentation in the context into concise and precise summary sequences.
Soft context compression approaches~\citep{ge2023context, chevalier2023adapting, bulatov2022recurrent, wang2024loma} offer a sound general-purpose compression framework. However, these approaches suffer from uncontrollable compression loss and lack of support for setting compression ratio when compressing tool documentation.
Therefore, we propose two strategies to improve context compression for tool-using language models.

The first strategy is selective compression, which mitigates compression loss on key information, i.e., names of tools and parameters.
We propose the construction of summary sequences in an interleaved format of compressed and uncompressed sub-sequences, wherein key information is preserved as original text tokens.
With less concern about losing key information, it is less challenging to condense summary sequences to a shorter length.

The second strategy is block compression which compresses documentation according to a fixed compression ratio instead of a fixed summary sequence length.
To achieve a controllable target compression ratio, we split the documentation into blocks according to the preset compression ratio, and then perform fixed-length compression separately.
In this manner, we no longer need to be concerned about length generalization or the wastage of summary tokens when dealing with documentation of a wide variety of lengths.

With these strategies, our approach starts from a pre-trained model, and then applies continual pre-training and fine-tuning pipeline to train the model to generate and leverage summary sequences.

We evaluate our approach on two tool-using benchmarks: API-Bank~\citep{li2023api} and APIBench~\citep{patil2023gorilla}.
Results show that under the compression ratio of up to 16x, our approach reaches comparable performance with the upper-bound baseline without compression.

We also explore the influence of our two proposed strategies on the same benchmarks.
Results show that selective compression significantly mitigates compression loss of key information, enabling a higher compression ratio.
Compared to overall compression, our block compression brings no additional compression loss.

Our primary contributions are as follows:
\begin{itemize}

    \item We introduce concise and precise context compression for tool-using language models, with strategies for minimizing key information loss under variable compression ratio.
    \item Our approach on two tool-using benchmarks demonstrates negligible performance loss under up to 16x compression ratio.
    \item We explore different combinations of training objectives and compression strategies, and provide a recipe for context compression training for tool-using language models.
\end{itemize}

\begin{figure*}[t!]
	\centering
	\includegraphics{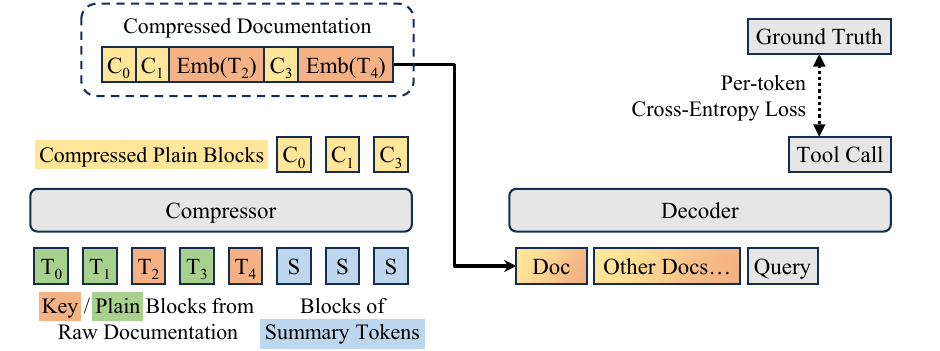}
	\caption{
		Overview of our method for tool documentation compression.
		When compressing a tool's documentation, we cut out the key information as key blocks (red) and chunk the rest into plain blocks (green).
		We use the concatenation of key blocks and compressed plain blocks (yellow) as the compressed documentation.
		We supervise the decoder output conditioned on compressed documentation to train the compressor and the decoder end-to-end.
	}
	\label{fig:method}
\end{figure*}

\section{Related Work}

\paragraph{Tool-Using Language Models}
In terms of flexibility, the tool-using capability of language models can be broadly categorized into two types: supporting limited pre-defined tools as built-in features, or enabling arbitrary tools by reading corresponding documentation in the context.
The first involves integrating predefined tools like search engines, calculators, and Python interpreters~\citep{cobbe2021training, nakano2021webgpt, komeili-etal-2022-internet, thoppilan2022lamda, gao2022pal, huang2022inner, yao2022react, cheng2023binding, schick2023toolformer}. The second more dynamically trains LMs to utilize tools by reading their documentation, expanding their application scope. This approach is exemplified by ChatGPT plugins~\citep{OpenAI_2023} and further developments in open-source LMs~\citep{yang2023gpt4tools, xu2023tool, patil2023gorilla, li2023api, qin2023toolllm, tang2023toolalpaca}, focusing on API usage and creating benchmarks for tool-using LMs.

Our study focuses on the second approach, where LMs generate structured function calls given tool documentation and instructions, illustrating practical application and flexibility in tool utilization.

\paragraph{Context Compression}

Context compression in large language models (LLMs) is pivotal for enhancing their efficiency. This research domain is mainly divided into soft summary and hard token pruning methods.

Soft summary methods, requiring pre-training or fine-tuning pre-trained language models as the compressor, yield high compression rates through condensed context representations. Prominent research in this area \citep{bulatov2022recurrent, ge2023context, wang2024loma} primarily utilizes memory tokens for context compression, yet often lacks focus on adjustable compression ratios.

Conversely, hard token pruning, exemplified by \citet{li2023compressing,jiang2023llmlingua, jiang2023longllmlingua}, entails the elimination of non-essential text by identifying and truncating non-informative sections.
Specifically, they measure the amount of information with self-information or the perplexity from language models.
Unlike soft summary methods, these methods leverage pure text protocol, thus easy to apply to black box models.

Our contribution enhances soft summary methods by integrating tool-use-specific strategies in LLMs and prioritizing controllable compression ratios. This approach, differing from works like \citet{chevalier2023adapting,ge2023context}, provides a customized solution for the specialized task in tool-using language models.

\section{Method}

Our approach uses two strategies to improve the basic soft context compression for tool-using language models.
In this paper, we name the language model which generates the output as the decoder, and refer to the model which generates the compressed sequence as the compressor.
Although the decoder and the compressor are the same model in our work, we differentiate them from each other in our description for clear logic.
We start with basic context compression, then introduce our two strategies, and finally describe our approach which integrates all of them.

\subsection{Basic Soft Context Compression}

As our starting point, basic soft context compression approaches aim to compress an arbitrary token sequence to a fixed length sequence of soft tokens~\citep{bulatov2022recurrent, chevalier2023adapting}, which we call overall compression.

To achieve this goal, a two-pass pipeline is performed to compress and then leverage the soft token sequence.

The first pass compresses the token sequence into a soft token sequence.
Assuming we have a token sequence $T = [t_0, t_1, ..., t_{L_T - 1}]$ to compress to $L_S$ soft tokens, we append a special summary token sequence $S = [s_0, s_1, ..., s_{L_S - 1}]$ to $T$ and input them to the compressor, obtaining the output hidden state
\[ H = \operatorname{Compressor}(\operatorname{Emb}(T) \Vert \operatorname{Emb}(S)) \text{.} \]
Then the hidden states $C = H[L_T, L_T + L_S)$ correspond to $S$ are used as the compressed soft token sequence, thus we have
\[ \operatorname{Compressed}(T) = H[L_T, L_T + L_S) \text{.} \]

The second pass leverages the soft token sequence as a soft prompt.
When generating the output for a context $T'$ conditioned on the compressed sequence $T$, $C$ is used as the alternate of $\operatorname{Emb}(T)$, thus we have the decoder output
\[ H' = \operatorname{Decoder}(C \Vert \operatorname{Emb}(T')) \text{.} \]

To train the compressor and decoder models, the language modeling training objective is applied to $H'$ as an indirect supervision since we cannot obtain the gold answer of soft tokens.

\subsection{Selective Compression Strategy}

The basic compression approaches have to face the challenge of compression loss. However, the compression loss is difficult to control. The higher the compression ratio, the greater the compression loss. Even if trained with reconstruction loss, the decoder can make mistakes when recovering words from compressed tokens, especially on rare words, just like human~\citep{ge2023context}.

This feature is harmful to tool-using language models, since key information loss such as misspelled names of tools or parameters will directly lead to failure.
Therefore, we propose the selective compression strategy as a more controllable approach to keep the key information despite the compression ratio, where the key information retains raw text tokens.

Given a tool documentation as the context $T$ to be compressed, we split it into disjoint sub-sequences $T_0, T_1, ..., T_{N_{\text{subseq}} - 1}$ whose union is $T$. Each $T_i$ is rather a key information sub-sequence (e.g., name of a parameter) or a sub-sequence that can be compressed (e.g., functionality description of the tool).
Following the notations in basic soft context compression, we have the compressed token sequence
\begin{align*}
	C &= \Vert_i C_i \text{ , where} \\
	C_i &= \left\{
	\begin{aligned}
		& \operatorname{Emb}(T_i) & T_i \text{ is key information} \\
		& \operatorname{Compressed}(T_i) & \text{otherwise} \\
	\end{aligned}
	\right. \text{.} \kern-\nulldelimiterspace & \hphantom{-1}
\end{align*}

\subsection{Block Compression Strategy}

A notable challenge in context compression for tool-using language models is the variability in documentation length. The basic compression methods typically compress the documentation into a uniform sequence of fixed length, denoted as $L_S$. This approach has limitations: for lengthy documentation, setting a small $L_S$ leads to a significant loss of information, adversely affecting performance. Conversely, a large $L_S$ hinders the effective compression of shorter documents, thereby reducing the compression ratio. Additionally, applying selective compression strategies, which divide the documentation into sub-sequences of varying lengths, intensifies these issues.

We believe it is a better approach to compress tool documentation according to a preset compression ratio $r$.
To realize this approach, we propose the block compression strategy to support the variable $r$ with fixed $L_S$.
The core idea is to chunk the sequence to be compressed into a variable number of blocks, each compressed to $L_S$ soft tokens.

We chunk the sequence to compress $T$ into $N_{\text{chunk}} = \lceil \frac{L_T}{r \times L_S} \rceil$ chunks $T_0, T_1, ..., T_{N_{\text{chunk}} - 1}$. Concatenating the compressed version of these chunks, we obtain the final compressed sequence
\[ C = \Vert_i \operatorname{Compressed}(T_i) \text{.} \]

Note that the last chunk is not always full, which will make the compressed sequence at most $L_S$ soft tokens longer than expected. Therefore, $L_S$ should be a small number.

\subsection{Concise and Precise Context Compression}

The aforementioned strategies offer a concise and precise approach to compress tool documentation for tool-using language models. As illustrated in Figure~\ref{fig:method}, our final method integrates both of them based on the basic soft context compression.

\paragraph{Combining Selective and Block Compression Strategies}

From the perspective of block compression, we can unify the two strategies in practice by regard key information sub-sequences as special blocks which we do not compress.
Specifically, we first split the key information sub-sequences (i.e., the key blocks), then chunk the other sub-sequences into blocks (i.e., the plain blocks).

Next, we input the blocks into the compressor.
To reach high efficiency and keep more context information, we compress all the blocks in one documentation in parallel.
As shown in the compression part of Figure~\ref{fig:method}, we append one block of summary tokens to the input sequence for each plain block, and obtain all the compressed blocks at once.

Finally, as described in the selective compression strategy, we concatenate all the blocks to form the compressed documentation, which is then used by the decoder.

\paragraph{Training Compressor and Decoder}

Following the basic soft context compression approaches~\citep{chevalier2023adapting, ge2023context}, we initialize the compressor and the decoder with a pre-trained language model, then jointly train them.
Actually, we use the same model as the compressor and the decoder, when we input summary tokens, it outputs compressed blocks, otherwise, it works as an ordinary language model.
The compressor and decoder need pre-training to acquire the capability of using soft tokens.
Different from existing approaches, our pre-training format changes due to the integration of our two strategies.
Specifically, we randomly chunk pre-training data as key blocks and plain blocks, and perform the same parallel compression manner as shown in Figure~\ref{fig:method}.

The training objective is language modeling, thus we apply the per-token cross-entropy loss on the decoder output. With the gradient propagated by the compressed documentation, the loss can supervise the decoder as well as the compressor in an end-to-end manner.

In addition,~\citet{ge2023context} propose to add an auxiliary loss of reconstructing the raw text $T$ from the compressed soft token sequence $C$, which agrees to our goal of keeping key information, so we take this idea into account in our method, implementing a variant of our approach with reconstruction loss.
In practice, we follow~\citet{ge2023context} to use the trainable soft prompt to switch the decoder between ordinary mode and reconstruction mode.

\section{Experiments}

\subsection{Basic Settings}

To evaluate our approach, we conduct experiments to train models with context compression and test the performance on tool-using benchmarks.
Our goal is to investigate the variations in model performance under different compression ratios and compression strategies.

\paragraph{Base Models}

In all of our experiments, we use the same base model as both the compressor and the decoder, and use LLaMA-7b~\citep{touvron2023llama} to initialize all the base models.
According to the manner of compression, we categorize base models into three cases:
\begin{itemize}
	\item No compression: this case corresponds to the fine-tuned LLaMA without context compression, which we regard as the upper-bound approach because it is not affected by compression loss.
	\item Overall context compression: these models act as our baselines, performing overall compression in the manner of basic soft context compression approaches.
	\item Selective context compression: these models benefit from our proposed selective compression strategy and demonstrate the performance of our approach.
\end{itemize}

In these cases, block compression is always enabled so that we can track the variation of model performance with a controllable compression ratio.

Note that we refer to existing compression approaches such as RMT~\citep{bulatov2022recurrent} and AutoCompressor~\citep{chevalier2023adapting} as basic compression, and the overall context compression case is equivalent to basic compression plus block compression.
Therefore, we study the influence of block compression strategy through extra analysis experiments in Section~\ref{sec:Effects of Block Compression Strategy} which provide fair comparison between our approach and these existing soft compression approaches.

\paragraph{Key Information for Selective Compression}

For general context, it is difficult to key information for selective compression.
However, when we focus on the documentation of tools, it becomes evident that the information that models must precisely retrieve is the name.
Therefore, we follow a simple but sound definition of key information as the names of tools and parameters.

\paragraph{Training Objectives}

We consider two training objectives in our experiments.
The first is language modeling in the supervision of the decoder output. It is the basic objective for all the base models including the no-compression case.
The second is the reconstruction objective, requiring the decoder to recover the raw text from compressed soft token sequences. Reconstruction is only for models with compression, acting as an auxiliary objective.
When the reconstruction objective is on, we use the sum of reconstruction and the language modeling loss as the final loss.

The reconstruction objective is from ICAE~\citep{ge2023context} which shares the same motivation of keeping the raw information and compatible with our approach.
From this perspective, ICAE is parallel to our approach. Therefore, we study the effects of the reconstruction loss as long as it is possible.
Specifically, switching the reconstruction objective on and off, each base model in the compression cases has two variants. We report and analysis results of both variants.

\subsection{Pre-training Compression Models}

Following existing soft compression approaches~\citep{chevalier2023adapting, ge2023context}, we pre-train base models on general corpus at first instead of directly fine-tuning them on downstream tasks.
The only exception is the upper-bound baseline, i.e., a fine-tuned LLaMA without context compression, for which pre-training is omitted because it does not need to acquire how to use soft tokens.

\paragraph{Dataset}

We use SlimPajama~\citep{cerebras2023slimpajama} as the pre-training dataset.
SlimPajama is a deduplicated version of RedPajama~\citep{together2023redpajama}, which is a community reproduction of the LLaMA~\citep{touvron2023llama} pre-training dataset.
To pre-train LLaMA with compression, we randomly sample data within 2k context length to construct a subset of 1B tokens.
Then, we train all the models with compression on the same subset for one epoch.

\paragraph{Compression Manner}

We maintain the compression manner during pre-training consistent with the manner in practice.
For overall compression models, we randomly select a prefix for each sample, with the length ranging from 0.5k to 1.5k, and then use the compressed prefix as input, with the remaining portions serving as output.
For selective compression models, we randomly mark sub-sequences of the prefix as key information according to a random proportion ranging from 0 to 1 for each sample.

Given the heavy computation cost of pre-training, to ensemble support for variable compression ratio in one model, we assign a random compression ratio ranging from 1 to 16 to each sample, and always set the length of the summary sequence as 2.

\paragraph{Instruction-Tuning}

Some tool-using language models including the official baseline of API-Bank are fine-tuned on instruction-tuned models~\citep{li2023api, tang2023toolalpaca, alpaca, vicuna2023}.
To unify the settings and make our base models prepared to efficiently adapt to downstream tasks, we follow them to instruction-tune the pre-trained compression models as the final base models.
Specifically, we use the ShareGPT dataset released by OpenChat~\citep{wang2023openchat} as the instruction-tuning dataset.
We trunk ShareGPT into a sequence length of 2k and train one epoch for all the models, with the other settings and training procedure the same as pre-training.

\subsection{Tool-Using Benchmarks}

\begin{table}[t]
	\centering
	\small
	\begin{tabular}{lccc}
		\toprule
		\bf Dataset & \tabincell{c}{\bf \#Samples \\ \bf Train / Test} & \bf \#Tools & \tabincell{c}{\bf Averaged \\ \bf Doc Length}  \\
		\midrule
		API-Bank & 6184 / 389 & 53 & 129  \\
		APIBench & 15034 / 1785 & 1726 & 356  \\
		\bottomrule
	\end{tabular}
	\caption{Statistics of tool-using benchmarks where documentation lengths are counted by the LLaMA tokenizer.}
	\label{tab:data_stat}
\end{table}

\begin{table*}[t]
	\centering
	\small
	\begin{tabular}{lcccccccc}
		\toprule
		& \multicolumn{4}{c}{\bf API-Bank} & \multicolumn{4}{c}{\bf APIBench}  \\
		\cmidrule(lr){2-5}\cmidrule(lr){6-9}
		\multicolumn{1}{c}{\bf Compression Ratio} & \bf 4 & \bf 8 & \bf 12 & \bf 16 & \bf 4 & \bf 8 & \bf 12 & \bf 16  \\
		\midrule
		No compression & \multicolumn{4}{c}{71.47} & \multicolumn{4}{c}{88.24}  \\
		\midrule
		Overall context compression & 68.12 & 67.10 & 64.52 & 61.70 & 88.18 & 88.12 & 85.15 & 85.71  \\
		\, w/ Reconstruction loss & 64.27 & 66.58 & 62.98 & 53.21 & 82.80 & 84.09 & 82.97 & 79.16  \\
		\hdashline\noalign{\vskip 0.5ex}
		Selective context compression & \bf 70.18 & \bf 72.75 & \bf 69.15 & \bf 72.49 & \bf 90.31 & 89.58 & 87.79 & \bf 89.13  \\
		\, w/ Reconstruction loss & 69.41 & 67.35 & 68.64 & 69.67 & 88.52 & \bf 89.75 & \bf 88.85 & 88.29  \\
		\bottomrule
	\end{tabular}
	\caption{Accuracy on the test set of two tool-using benchmarks. The performance of selective compression is seldom affected by the compression ratio, however, the performance of overall compression noticeably decays as the compression ratio increases.}
	\label{tab:exp_sel_vs_ova}
\end{table*}

We evaluate our approach on tool-using benchmarks API-Bank~\citep{li2023api} and APIBench~\citep{patil2023gorilla}.
Both of them can be modeled as a standard case in the decoder part of Figure~\ref{fig:method}, in which tool-using language models accept one or more tool documentation as well as the user query as input and then output the tool call.

\paragraph{API-Bank}

The dataset consists of multi-turn dialogues, where the user can ask the model to call external APIs.
Each tool documentation is a JSON dictionary as exemplified in Figure~\ref{fig:task}.
We use the \texttt{level1-api} subset of API-Bank in our experiments, where each of the user queries is a dialog history ending with an instruction to use a tool.

\paragraph{APIBench}

The dataset simulates a scene where an automated agent finds a suitable model on a platform (e.g., Hugging Face Hub) to fulfill the user's query.
Therefore, the input contains a query and the model card of candidate models as the tool documentation, and the output is an API call to drive the model.
However, the dataset is originally for testing retrieval-augmented tool-using language models.
Specifically, only the top-ranked candidate model retrieved is provided to the decoder, and it cannot be guaranteed that the retrieval is correct, thus introducing the possibility of cascading errors.
To weaken the impact of the retrieval module, we use the BM25 retrieval module in the official codebase to extend the number of candidates to up to 5, and make sure the correct answer is within. We shuffle the order of documentation to avoid potential position bias.
The original dataset contains three subsets (i.e., \texttt{huggingface}, \texttt{tensorflowhub}, and \texttt{torchhub}), we use their union as a whole dataset.

\paragraph{Fine-tuning on Downstream Tasks}

We use the official training and test set for both of the datasets, whose statistics are shown in Table~\ref{tab:data_stat}.
As shown in Figure~\ref{fig:method}, we always compress different tool documentation separately, then concatenate all the compressed documentation before inputting them to the decoder.
To ensure the consistency between training and testing, we train a separate model for each combination of the compression ratio, the base model, and the dataset.
We train all the models on the corresponding dataset for two epochs at once.

\paragraph{Metric}

Both of the benchmarks use the accuracy of API calls as the metric.
API-Bank provides a local sandbox to run the APIs, and check the running result to judge whether the API call is correct or not.
APIBench checks the answer through AST matching, without running the API.
We follow the official test approaches for both of the benchmarks.

\subsection{Tool-Using Evaluation}

Table~\ref{tab:exp_sel_vs_ova} demonstrates the performance of three cases of base models.
In general, selective context compression outperforms overall context compression, reaching similar or even higher performance compared to the no-compression case.
Also, we find that adding reconstruction loss can be harmful to performance, and selective context compression can close the performance gap caused by reconstruction loss.

In these two benchmarks, trends of performance according to the compression ratio differ.
Table~\ref{tab:data_stat} shows that documentation in API-Bank is more concise than those in APIBench, thus intuitively harder to compress.
Results show that the performance of overall context compression noticeably decays when the compression ratio becomes higher, which supports this intuition.
On the other hand, only the weakest base model, which is overall context compression with reconstruction loss, shows obvious performance degradation on APIBench.
This phenomenon suggests that APIBench is much easier than API-Bank, having the potential to keep satisfactory performance under a higher compression ratio.

\begin{table}[t]
	\setlength\tabcolsep{3.5pt}
	\centering
	\small
	\begin{tabular}{lcccccccc}
		\toprule
		& \multicolumn{4}{c}{\bf API-Bank} & \multicolumn{4}{c}{\bf APIBench}  \\
		\cmidrule(lr){2-5}\cmidrule(lr){6-9}
		\multicolumn{1}{c}{\bf Comp. Ratio} & \bf 4 & \bf 8 & \bf 12 & \bf 16 & \bf 4 & \bf 8 & \bf 12 & \bf 16  \\
		\midrule
		No comp. & \multicolumn{4}{c}{21} & \multicolumn{4}{c}{101}  \\
		\midrule
		Overall comp. & 39 & 31 & 41 & 51 & 106 & 107 & 128 & 129  \\
		\, w/ Rec. loss & 53 & 41 & 45 & 93 & 178 & 168 & 185 & 209  \\
		\hdashline\noalign{\vskip 0.5ex}
		Selective comp. & 30 & \bf 21 & \bf 33 & \bf 21 & \bf 83 & 86 & 107 & \bf 85  \\
		\, w/ Rec. loss & \bf 24 & 36 & \bf 33 & 27 & 103 & \bf 85 & \bf 99 & 111  \\
		\bottomrule
	\end{tabular}
	\caption{Number of name errors corresponds to Table~\ref{tab:exp_sel_vs_ova}, lower is better.}
	\label{tab:exp_sel_vs_ova_name_error}
\end{table}

\begin{table*}[t]
	\setlength\tabcolsep{5pt}
	\centering
	\small
	\begin{tabular}{lcccccccccc}
		\toprule
		\multirow{2}{*}[-1ex]{\bf Approach} & \multirow{2}{*}[-1ex]{\tabincell{c}{\bf Pre-training \\ \bf Strategy}} & \multirow{2}{*}[-1ex]{\tabincell{c}{\bf Fine-tuning \\ \bf Strategy}} & \multicolumn{4}{c}{\bf API-Bank} & \multicolumn{4}{c}{\bf APIBench}  \\
		\cmidrule(lr){4-7}\cmidrule(lr){8-11}
		&  &  & \bf 4$\bf \times$ & \bf 8$\bf \times$ & \bf 12$\bf \times$ & \bf 16$\bf \times$ & \bf 4$\bf \times$ & \bf 8$\bf \times$ & \bf 12$\bf \times$ & \bf 16$\bf \times$  \\
		\midrule
		No compression & n/a & n/a & \multicolumn{4}{c}{71.47} & \multicolumn{4}{c}{88.24}  \\
		\midrule
		\multirow{4}{*}{Compression} & Overall & Overall & 68.12 & 67.10 & 64.52 & 61.70 & 88.18 & 88.12 & 85.15 & 85.71  \\
		& Selective & Overall & 67.35 & 60.15 & 69.67 & 67.10 & 88.07 & 88.12 & 89.19 & 86.11  \\
		& Overall & Selective & \bf 70.44 & 69.15 & \bf 70.18 & 68.38 & 85.49 & 88.63 & \bf 89.75 & \bf 89.41  \\
		& Selective & Selective & 70.18 & \bf 72.75 & 69.15 & \bf 72.49 & \bf 90.31 & \bf 89.58 & 87.79 & 89.13  \\
		\midrule
		\multirow{4}{*}{\tabincell{l}{Compression \\ \, w/ Reconstruction loss}} & Overall & Overall & 64.27 & 66.58 & 62.98 & 53.21 & 82.80 & 84.09 & 82.97 & 79.16  \\
		& Selective & Overall & 66.84 & 65.30 & 65.30 & 65.81 & 86.05 & 87.34 & 83.64 & 83.03  \\
		& Overall & Selective & 68.89 & \bf 69.15 & 66.84 & 65.04 & 87.00 & 87.96 & 85.49 & 85.94  \\
		& Selective & Selective & \bf 69.41 & 67.35 & \bf 68.64 & \bf 69.67 & \bf 88.52 & \bf 89.75 & \bf 88.85 & \bf 88.29  \\
		\bottomrule
	\end{tabular}
	\caption{Exploration on the effects of selective compression in different training stages. Compression models can adapt to selective compression with fine-tuning. The best combination is to consistently train with selective compression. 4$\times$ to 16$\times$ represents the compression ratios.}
	\label{tab:exp_sel}
\end{table*}

Furthermore, we dive into the error cases to explore whether selective context compression keeps key information or not.
Under our intuitive definition of key information as the tool names and parameter names, we count the number of error cases caused by name error, i.e., the model predicts a wrong name of APIs or parameters.

From results aggregated in Table~\ref{tab:exp_sel_vs_ova_name_error}, we find selective compression models make fewer mistakes in name errors.
On APIBench, this phenomenon is more obvious. Although overall compression can reach comparable performance with the no compression baseline, the number of name errors grows with the compression ratio.
In contrast, selective compression keeps the number of name errors even lower than baseline, less affected by the compression ratio.

Another interesting finding is that the overall compression base model with reconstruction loss produces even more name errors, suggesting that without priors it is hard for the model to realize the importance of names.

\subsection{Effects of Block Compression Strategy}

\label{sec:Effects of Block Compression Strategy}

\begin{table}[t]
	\setlength\tabcolsep{4pt}
	\centering
	\small
	\begin{tabular}{lcc}
		\toprule
		& \multicolumn{1}{c}{\bf API-Bank} & \multicolumn{1}{c}{\bf APIBench}  \\
		\midrule
		No compression & 71.47 & 88.24  \\
		\midrule
		\multicolumn{3}{l}{\it Compress each documentation separately}  \\
		\midrule
		Basic context compression & 56.81 & 82.52  \\
		\, w/ Block context compression & \bf 69.92 & \bf 85.88  \\
		\midrule
		\multicolumn{3}{l}{\it Compress all documentation as a whole}  \\
		\midrule
		Basic context compression & 51.41 & 80.39  \\
		\, w/ Block context compression & \bf 64.52 & \bf 81.85  \\
		\bottomrule
	\end{tabular}
	\caption{When compressing documentation to 50 soft tokens, the proposed block compression greatly benefits existing basic compression approach, i.e., applying RMT or AutoCompressor directly.}
	\label{tab:exp_blk}
\end{table}

Since supporting of controllable compression ratio relies on block compression, the methodology of observing the performance in different compression ratios does not work when studying block compression itself.
To make a fair comparison between cases with and without block compression, we step back to the basic soft context compression.
This section also plays the role of providing fair comparison between existing soft compression approaches, namely RMT~\citep{bulatov2022recurrent} and AutoCompressor~\citep{chevalier2023adapting}, and the proposed block comporession strategy.

Actually, the overall context compression base model is a basic context compression model with the integration of block compression.
Therefore, we train a new basic context compression base model without block compression with the same data and pre-training-fine-tuning pipeline with the overall context compression base model.
We thereby can analyze the influence of block compression through comparing the performance of these two base models under the same length of the compressed summary token sequence.
We follow AutoCompressor~\citep{chevalier2023adapting} to set the length of summary sequences to 50.

Table~\ref{tab:exp_blk} demonstrates experiment results.
Apart from the plain setting of separately compressing each documentation, we add another setting to observe the performance under a higher compression ratio, where we regard the concatenation of multiple documentation as a whole.
With either setting, block compression significantly improves basic context compression.

\subsection{Effects of Selective Compression Strategy}

To study the effects of selective compression strategy and the necessity of selective strategy in pre-training, we evaluate all four cases of strategy combination during pre-training and fine-tuning.
Results are shown in Table~\ref{tab:exp_sel}, which can be seen as an extended version of Table~\ref{tab:exp_sel_vs_ova}.

We find that introducing selective compression benefits the performance even if the final compression manner is overall compression.
Moreover, the base model can adapt to selective compression through fine-tuning, though having a performance gap to the best combination.
To reach the best performance, the model should use selective compression from the beginning to the end.
The conclusion is supported by Table~\ref{tab:exp_sel} both with and without the reconstruction loss.

\section{Conclusion}

In this work, we propose an approach to compress tool documentation into concise and precise summary sequences.
There are two main challenges in achieving our gold. First, context compression approaches suffer from uncontrollable compression loss, leading to key information loss. Second, existing approaches cannot generate summary sequences of variable length, and thus cannot support preset compression ratio.
We propose two compression strategies to deal with these challenges. To avoid key information loss, we propose the selective compression strategy which allows the key information to be retained as raw text tokens. To support preset compression ratios, we propose block compression which chunks the full sequence to be compressed into small blocks to realize variable length compression upon fixed length compression model.

We evaluate our approach in two tool-using benchmarks. Results show that our method can reach at least comparable performance to the baseline without compression, while achieving up to 16x compression ratio.
Furthermore, we explore the effects of our two compression strategies, and provide a training recipe of context compression for tool-using language models as a result.

\section*{Limitations}

The main limitation of our work is that the proposed strategies rely on human priors.
Specifically, the definition of key information and the preset compression ratio need to be tuned according to the actual task.

Another limitation is that our compression approach needs to pre-train the model, which means the computation cost is relatively high, and the approach is not suitable for black box models. Although this is a universal problem of soft context compression approaches, it still hinders the flexibility of our approach.

\section*{Acknowledgements}

We gratefully acknowledge the support of the National Natural Science Foundation of China (NSFC) via grants 62236004 and 62206078.

\bibliography{custom}

\appendix

\section{Additional Results}

\subsection{The Necessity of Non-key Information}

Although we use open datasets from existing work, there not exists evidence supporting the necessity of non-key information such as the descriptions. Theoretically, the model may guess the functionality of a tool only using the names of the APIs and parameters.

Therefore, we conduct experiments on the case that non-key information is deleted. In other words, the tool documentation consist of almost only the names.

\begin{table}[h]
	\setlength\tabcolsep{4pt}
	\centering
	\small
	\begin{tabular}{cccc}
		\toprule
		\bf Dataset & \bf Fine-tuned Input & \bf Test Input & \bf Accuracy  \\
		\midrule
            \multirow{4}{*}{API-Bank} & Full & Full & \bf 71.47  \\
             & Full & Key-only & 53.21  \\
             & Key-only & Full & 59.38  \\
             & Key-only & Key-only & 40.62  \\
		\midrule
            \multirow{4}{*}{APIBench} & Full & Full & \bf 88.24  \\
             & Full & Key-only & 67.45  \\
             & Key-only & Full & 27.51  \\
             & Key-only & Key-only & 83.75  \\
		\bottomrule
	\end{tabular}
	\caption{Deleting non-key information significantly hurts the performance, which supports the necessity of non-key information.}
	\label{tab:exp_del_nonkey}
\end{table}

The results are shown in Table~\ref{tab:exp_del_nonkey}. To understand the results, please note that tools in the API-Bank test set are unseen during training. In contrast, APIBench shares the same tool library during training and testing. Thus, when fine-tuned and tested with the same type of input, the model has the chance to memorize the tools in APIBench. However, a performance gap also exists in this case. To summarize, we can conclude that the non-key information is necessary.

\subsection{Comparison with Hard Summarization}

Apart from soft compression which we use, another way to achieve tool documentation compression is hard compression, such as prompting ChatGPT to summarize the documentation.
However, hard tokens are able to carry less information in the same context length, which is the reason we consider soft compression at first.

To illustrate the manner of hard compression, we use GPT-4 Turbo to compress the non-key part of the tool documentation, which is a very close setting to our main experiment despite we use ChatGPT as the compressor.

We explain the experiment details on API-Bank, which are highly similar to the case on APIBench. We compress all \texttt{description} fields since other non-key fields like datatype consist of very short text. Specifically, we use the following prompt and switch on the JSON-only output mode:

\begin{Verbatim}[breaklines=true]
Here is an API document.


{{json.dumps(the_documentation_dict, ensure_ascii=False)}}


Replace each "description" field with a brief summary and keep the other parts as is. The summary should remove redundancy and express the text as concisely as possible, ensuring that allkey information are preserved. Only output a single json without the quote block.
\end{Verbatim}

Also, we carefully check the output and leverage regenerating to ensure any other field is kept as is.

To provide better intuition, we give an example as follows (formatted for easy reading).

\begin{Verbatim}[breaklines=true]
# the raw doc
{
    "name": "Translate",
    "description": "Translate the text to the target language.",
    "input_parameters": {
        "src": {
            "type": "str",
            "description": "The text to be translated."
        },
        "src_lang": {
            "type": "str",
            "description": "[Optional] The source language to translate from. Default is auto."
        },
        "tgt_lang": {
            "type": "str",
            "description": "[Optional] The target language to translate to. Default is english/en."
        }
    },
    "output_parameters": {
        "translated_text": {
            "type": "str",
            "description": "The translated text."
        }
    }
}

# doc after GPT-4 summarization
{
    "name": "Translate",
    "description": "Translates text to a specified language.",
    "input_parameters": {
        "src": {
            "type": "str",
            "description": "Text for translation."
        },
        "src_lang": {
            "type": "str",
            "description": "Source language (auto by default)."
        },
        "tgt_lang": {
            "type": "str",
            "description": "Target language (English by default)."
        }
    },
    "output_parameters": {
        "translated_text": {
            "type": "str",
            "description": "Resulting translation."
        }
    }
}
\end{Verbatim}

We can see from the example that plain text cannot compress concise input with a high compression ratio, while soft compression has the chance to compress a short description into a single soft token at an over 10$\times$ compression ratio.

The only difference on APIBench is that we compress the \texttt{example\_code} as well as the \texttt{description} field. The other fields are always very short.

Next, we fine-tune using the baseline setting in the paper, namely fine-tuning a LLaMA-7b model fine-tuned on ShareGPT. To avoid inconsistency between training and testing, we also evaluate the case where the training data are also compressed.

\begin{table}[h]
	\setlength\tabcolsep{4pt}
	\centering
	\small
	\begin{tabular}{cccc}
		\toprule
		\bf Dataset & \bf Fine-tuned Input & \bf Test Input & \bf Accuracy  \\
		\midrule
            \multirow{4}{*}{API-Bank} & Raw & Raw & \bf 71.47  \\
            & Raw & GPT-4 sum & 67.87  \\
            & GPT-4 sum & Raw & 55.53  \\
            & GPT-4 sum & GPT-4 sum & 52.70  \\
		\midrule
            \multirow{4}{*}{APIBench} & Raw & Raw & \bf 88.24  \\
            & Raw & GPT-4 sum & 87.79  \\
            & GPT-4 sum & Raw & 86.27  \\
            & GPT-4 sum & GPT-4 sum & 86.44  \\
		\bottomrule
	\end{tabular}
	\caption{GPT-4 Turbo hard summarization perform worse in accuracy than our approach.}
	\label{tab:exp_gpt4_acc}
\end{table}

\begin{table}[h]
	\setlength\tabcolsep{4pt}
	\centering
	\small
	\begin{tabular}{ccc}
		\toprule
		\bf Approach & \bf Dataset & \bf \tabincell{c}{Achieved \\ Compression Ratio}  \\
		\midrule
            \multirow{2}{*}{GPT-4 sum} & API-Bank & 1.39$\times$  \\
             & APIBench & 1.42$\times$  \\
		\midrule
            Ours & \tabincell{c}{(nearly) \\ dataset agnostic} & \tabincell{c}{configurable, \\ up to 16$\times$ at least}  \\
		\bottomrule
	\end{tabular}
	\caption{GPT-4 Turbo hard summarization achieves lower compression ratios than our approach.}
	\label{tab:exp_gpt4_comp_ratio}
\end{table}

Table~\ref{tab:exp_gpt4_acc} lists the accuracy and Table~\ref{tab:exp_gpt4_comp_ratio} lists the averaged compression ratio over the datasets, where we find GPT-4 summarization is less efficient. Tested with GPT-4 summarized input, the baseline model achieves lower performance. When trained with GPT-4 summarized data, the performance goes even lower. Note that the official training set of API-Bank uses a dedicated set of tool documentation which are model generated thus far more messy and harder to summarize than the test data. In contrast, APIBench uses the same set of tool documentation in the training set and the test set. These results imply a possible flaw that untrained hard summarization may introduce information loss or error on tool-using tasks.

Please note that our work mainly focuses on developing an efficient approach to compressing tool documentation based on the soft context compression framework, instead of a comparison of soft/hard compression approaches on specific tasks. Therefore, the results of hard summary baselines act as additional data to help readers better understand our motivation and better illustrate the differences between soft/hard compression approaches.
Anyway, the performance of hard summarization/compression has a limited relation with the integrity of our work.

\end{document}